\documentclass{ifacconf}

\usepackage{graphicx}      
\usepackage{natbib}        

\usepackage{mathtools} 

\usepackage{threeparttable}

\usepackage{amsmath,amssymb,amsfonts}
\usepackage{algorithmic}
\usepackage{graphicx}
\usepackage{textcomp}
\usepackage{xcolor}

\usepackage{cuted}   
\usepackage{capt-of} 
\usepackage{textcase}

\usepackage{stfloats}

\usepackage{url}

\newcommand{\twolinetauB}[2]{%
  \noindent\(\displaystyle #1 \)\\[0.35em]%
  \noindent\(\left[\!\left[
    \begin{array}{@{}l@{}}
      #2
    \end{array}
  \right]\!\right]\)%
}

\usepackage{tabularx} 

\usepackage{booktabs}
\usepackage{placeins}
\usepackage{siunitx}
\DeclareSIUnit{\degree}{\ensuremath{^\circ}}

\IfFileExists{wasysym.sty}{
  \usepackage{wasysym}             
  \newcommand{\Cfull}{\CIRCLE}     
  \newcommand{\Chalf}{\LEFTcircle} 
  \newcommand{\Cempty}{\Circle}    
}{
  \usepackage{tikz}
  \newcommand{\Cfull}{\tikz[baseline=-0.6ex]\fill (0,0) circle (0.6ex);}
  \newcommand{\Cempty}{\tikz[baseline=-0.6ex]\draw (0,0) circle (0.6ex);}
  \newcommand{\Chalf}{\tikz[baseline=-0.6ex]{%
    \draw (0,0) circle (0.6ex);
    \clip (-0.6ex,-0.6ex) rectangle (0,0.6ex); 
    \fill (0,0) circle (0.6ex);
  }}
}

\usepackage{pifont}   

\usepackage{stfloats}

\usepackage{siunitx}
\makeatletter

\newcommand{\Rmnum}[1]{\expandafter\@slowromancap\romannumeral #1@}
\makeatother

\usepackage{booktabs}

\usepackage{enumitem} 

\def\BibTeX{{\rm B\kern-.05em{\sc i\kern-.025em b}\kern-.08em
    T\kern-.1667em\lower.7ex\hbox{E}\kern-.125emX}}
    
\begin{document}
\begin{frontmatter}


\title{Towards Logic-Aware Manipulation: A Knowledge Primitive for VLM-Based Assistants in Smart Manufacturing\thanksref{footnoteinfo}} 

\thanks[footnoteinfo]{Acknowledgment to Guangzhou-HKUST(GZ) Joint Funding Program (2025A03J3658), and National Natural Science Foundation of China (72231005).}

\author[First]{Suchang Chen} 
\author[First]{Daqiang Guo} 

\address[First]{The Hong Kong University of Science and Technology (Guangzhou), 
    No.1 Du Xue Rd, Nansha District, Guangzhou, Guangdong, China (e-mail: schen522@connect.hkust-gz.edu.cn, daqiangguo@hkust-gz.edu.cn)}

\begin{abstract}
Existing pipelines for vision-language models (VLMs) in robotic manipulation prioritize broad semantic generalization from images and language, but typically omit execution-critical parameters required for contact-rich actions in manufacturing cells. We formalize an object-centric manipulation-logic schema, serialized as an eight-field tuple $\tau$, which exposes object, interface, trajectory, tolerance, and force/impedance information as a first-class knowledge signal between human operators, VLM-based assistants, and robot controllers. We instantiate $\tau$ and a small knowledge base (KB) on a 3D-printer spool-removal task in a collaborative cell, and analyze $\tau$-conditioned VLM planning using plan-quality metrics adapted from recent VLM/LLM planning benchmarks, while demonstrating how the same schema supports taxonomy-tagged data augmentation at training time and logic-aware retrieval-augmented prompting at test time as a building block for assistant systems in smart manufacturing enterprises.
\end{abstract}


\end{frontmatter}

\section{Introduction}\label{sec:intro}

Vision-language models (VLMs) have moved robotic manipulation beyond brittle, task-specific pipelines: language-conditioned policies execute diverse tabletop and deformable tasks, multimodal prompts enable systematic generalization, and cross-robot corpora support transfer to new embodiments\citep{driess_palm-e_2023, brohan_rt-1_2023, kim_openvla_2025}. Yet these advances emphasize \emph{what} to act on and \emph{where} to move rather than the execution logic that makes first contact succeed \citep{shridhar_perceiver-actor_2023,jiang_vima_2023}. In smart manufacturing, failure often arises at contact: appearance-driven, generic policies lack interface mechanism, contact modality, trajectory shaping, precision/tolerance bands, and force/impedance, so robots frequently misalign, bind, slip, or violate limits on the first attempt.\citep{stone_open-world_2023} This under-specification is well documented for contact-rich, precision-sensitive operations where the goal is visually obvious but the procedure must satisfy simultaneous motion and force constraints \citep{raibert_hybrid_1981,hogan_impedance_1984}.


We propose an object-centric manipulation-logic schema that elevates these details to a first-class signal for VLMs. The schema covers object parts and interfaces, motion primitives and directionality, trajectory profiles and timing, precision/tolerance bands, and force/impedance with feedback criteria. Prior evidence shows that structured metadata and language conditioning improve generalization, while diverse pretraining and explicit affordance structure narrow the gap between semantics and executable behavior \citep{shridhar_cliport_2022,nair_r3m_2023}.

\textbf{The contributions of this paper are threefold.}
\vspace{-2pt}
\begin{itemize}
  \setlength\itemsep{0pt}\setlength\parskip{0pt}\setlength\parsep{0pt}
\item \textbf{Schema.} An object-centric manipulation-logic schema with minimal, unit-bearing slots for interface, preconditions, contact modality, motion primitive, trajectory, tolerances, and impedance/safety limits, designed for VLM I/O.
\item \textbf{System and dual-use knowledge base (KB).} A logic-aware system where a compact knowledge base of $\tau$ entries supports (i) schema-tagged augmentation of documentation, demonstrations, or logs at training time, and (ii) retrieval-based conditioning of prompts and controllers at test time.
\item \textbf{Plan-level evidence and metrics.} A 3D-printer spool-removal case study in which $\tau$-conditioned prompts change VLM/LLM planning behavior under standard plan-quality metrics (completeness, order validity, safety/constraint coverage, parameter specificity, plan length), plus defined train-time and deployment metrics for future $\tau$-aware VLM training and hardware evaluation.
\end{itemize}
\vspace{-4pt}

\providecommand{\Cfull}{\ensuremath{\bullet}}
\providecommand{\Chalf}{\ensuremath{\circ}}
\providecommand{\Cempty}{\ensuremath{\cdot}}


\begin{table*}[!b]
\caption{\NoCaseChange{Family-level coverage of }$\tau$}
\label{tab:tau-gap}
\centering
\footnotesize
\setlength{\tabcolsep}{6pt}
\renewcommand{\arraystretch}{1.1}

\begin{threeparttable}
\footnotesize
\setlength{\tabcolsep}{6pt}
\renewcommand{\arraystretch}{1.1}

\begin{tabular}{p{0.52\textwidth}cccccccc}
\toprule
\textbf{Family of research (representative works)} & \textbf{obj} & \textbf{iface} & \textbf{pre} & \textbf{contact} & \textbf{prim} & \textbf{traj} & \textbf{tol} & \textbf{dyn} \\
\midrule
VLM planners (CLIPort\citep{shridhar_cliport_2022},  VIMA\citep{jiang_vima_2023}, PerAct\citep{shridhar_perceiver-actor_2023}, VoxPoser\citep{huang_voxposer_2023}, Instruct2Act\citep{huang_instruct2act_2023})
  & \Cfull & \Cempty & \Chalf & \Cempty & \Chalf & \Chalf & \Cempty & \Cempty \\
Generalist VLA (PaLM-E\citep{driess_palm-e_2023}, RT-1\citep{brohan_rt-1_2023}, OpenVLA\citep{kim_openvla_2025})
  & \Cfull & \Cempty & \Chalf & \Cempty & \Chalf & \Cempty & \Cempty & \Cempty \\
Affordance grounding (Open-vocabulary 3D affordance\citep{nguyen_open-vocabulary_2023}, OVAL-Prompt\citep{tong_oval-prompt_2024}, AffordanceLLM\citep{qian_affordancellm_2024}, Chain-of-Affordance\citep{li_improving_2024}, ReKep\citep{huang_rekep_2025})
  & \Cfull & \Chalf & \Cempty & \Chalf & \Chalf & \Chalf & \Cempty & \Cempty \\
Symbolic knowledge bases (FOON\citep{paulius_functional_2018})
  & \Cfull & \Chalf & \Cfull & \Chalf & \Cfull & \Chalf & \Cempty & \Cempty \\
Manual parsing from manuals (Manual2Skill\citep{tie_manual2skill_2025})
  & \Cfull & \Chalf & \Cfull & \Chalf & \Chalf & \Cfull & \Chalf & \Cempty \\
Retrieval-augmented planning (ExRAP\citep{yoo_exploratory_2024}, P-RAG\citep{xu_p-rag_2024}, RoboMP$^{2}$ \citep{lv_robomp2_2024})
  & \Chalf & \Chalf & \Cfull & \Chalf & \Chalf & \Chalf & \Cempty & \Cempty \\
Contact-aware learned control (Diffusion Policy\citep{chi_diffusion_2025}; Impedance IL\citep{zhou_variable_2025})
  & \Cempty & \Cempty & \Cempty & \Cfull & \Chalf & \Cfull & \Chalf & \Cfull \\
Classical contact control (Hybrid position/force control\citep{raibert_hybrid_1981}; Impedance control\citep{hogan_impedance_1984})
  & \Cempty & \Cempty & \Cempty & \Cfull & \Cempty & \Chalf & \Cempty & \Cfull \\
Industrial case studies(IKEA Chair Assembly \citep{suarez-ruiz_can_2018}, Peg-in-hole \citep{chen_robust_2025})
  & \Chalf & \Chalf & \Chalf & \Cfull & \Cfull & \Cfull & \Cfull & \Cfull \\
Multimodal sensing datasets (DIGIT tactile \citep{lambeta_digit_2020}; Touch--Language--Vision\citep{cheng_touch100k_2025} )
  & \Chalf & \Chalf & \Cempty & \Cfull & \Cempty & \Cempty & \Cempty & \Chalf \\
Contact-rich assembly datasets (REASSEMBLE\citep{sliwowski_demonstrating_2025})
  & \Cfull & \Chalf & \Chalf & \Cfull & \Cfull & \Cfull & \Chalf & \Cfull \\
\bottomrule
\end{tabular}

\begin{tablenotes}[para,flushleft]
\footnotesize
\item \NoCaseChange{Family-level coverage of the interaction tuple $\tau$ (obj, iface, pre, contact, prim, traj, tol, dyn).} Cells use circle marks to indicate degree of treatment: \protect\Cfull\ full/explicit coverage, \protect\Chalf\ partial or implicit coverage, \protect\Cempty\ not covered. Representative works for each family are listed in parentheses.
\end{tablenotes}

\end{threeparttable}
\end{table*}

\section{Background and Motivation}\label{sec:bg}

Vision–language models (VLMs) broaden semantic generalization for manipulation but often miss execution details that determine contact success: approach strategy, force/torque application, dwell timing, and precision/tolerance bands. Classic control shows appearance-conditioned instructions are insufficient: contact demands coordinated regulation of motion and interaction forces via impedance or hybrid position–force control, not a single nominal trajectory \citep{hogan_impedance_1984,raibert_hybrid_1981}. In collaborative settings, compliant/admittance control further treats precision bands and interaction dynamics as first-class parameters \citep{zhu_compliant_2025}.

Two complementary threads indicate how explicit structure moves VLMs from \emph{what} to \emph{how}. First, structured metadata and affordance-centric supervision map parts to operations, tightening the link between semantics and executable behavior and improving generalization \citep{tong_oval-prompt_2024,qian_affordancellm_2024}. Second, logic-aware prompting with retrieval supplies per-instance procedures at runtime; retrieval-augmented planning and memory-augmented task reasoning show gains when procedural constraints are injected before execution, especially for long-horizon tasks \citep{yoo_exploratory_2024,fan_vision-language-guided_2024}.

Our focus is the minimal schema that carries the execution variables those methods typically leave implicit. Functional graphs capture object–action–state structure\citep{paulius_functional_2018}; language-conditioned imitation and representation learning provide semantic and spatial grounding with data efficiency \citep{shridhar_cliport_2022,nair_r3m_2023}; multimodal prompting and program synthesis support generalization \citep{jiang_vima_2023,huang_voxposer_2023,shridhar_perceiver-actor_2023,huang_instruct2act_2023}; open-vocabulary part/affordance methods and reasoning datasets localize actionable parts \citep{qian_affordancellm_2024,li_improving_2024, huang_rekep_2025}; retrieval-augmented planning and memory-augmented reasoning improve long-horizon performance \citep{fan_vision-language-guided_2024, xu_p-rag_2024, lv_robomp2_2024,yoo_exploratory_2024}; task-and-motion planning accepts symbolic/continuous constraints \citep{garrett_pddlstream_2020}; skill-grounded planners enhance feasibility \citep{ahn_as_2023,huang_inner_2023}; and generalist corpora broaden capability \citep{brohan_rt-1_2023}. Coverage of the schema components (defined in Section~\ref{sec:schema}) across these families is summarized in Tab.~\ref{tab:tau-gap}. Collectively, these lines of work motivate a minimal, unit-bearing schema that (i) declares contact-aware parameters, (ii) serves as a retrieval and supervision target linking perception, language, and action, and (iii) provides a runtime check to tolerances and interaction dynamics. Prior families typically leave tolerances and dynamics implicit, decouple training-time annotation from inference-time retrieval, and rarely expose semantics or numeric limits as checkable execution variables.

\begin{figure*}[b!]
  \centering
  \includegraphics[width=13.6 cm]{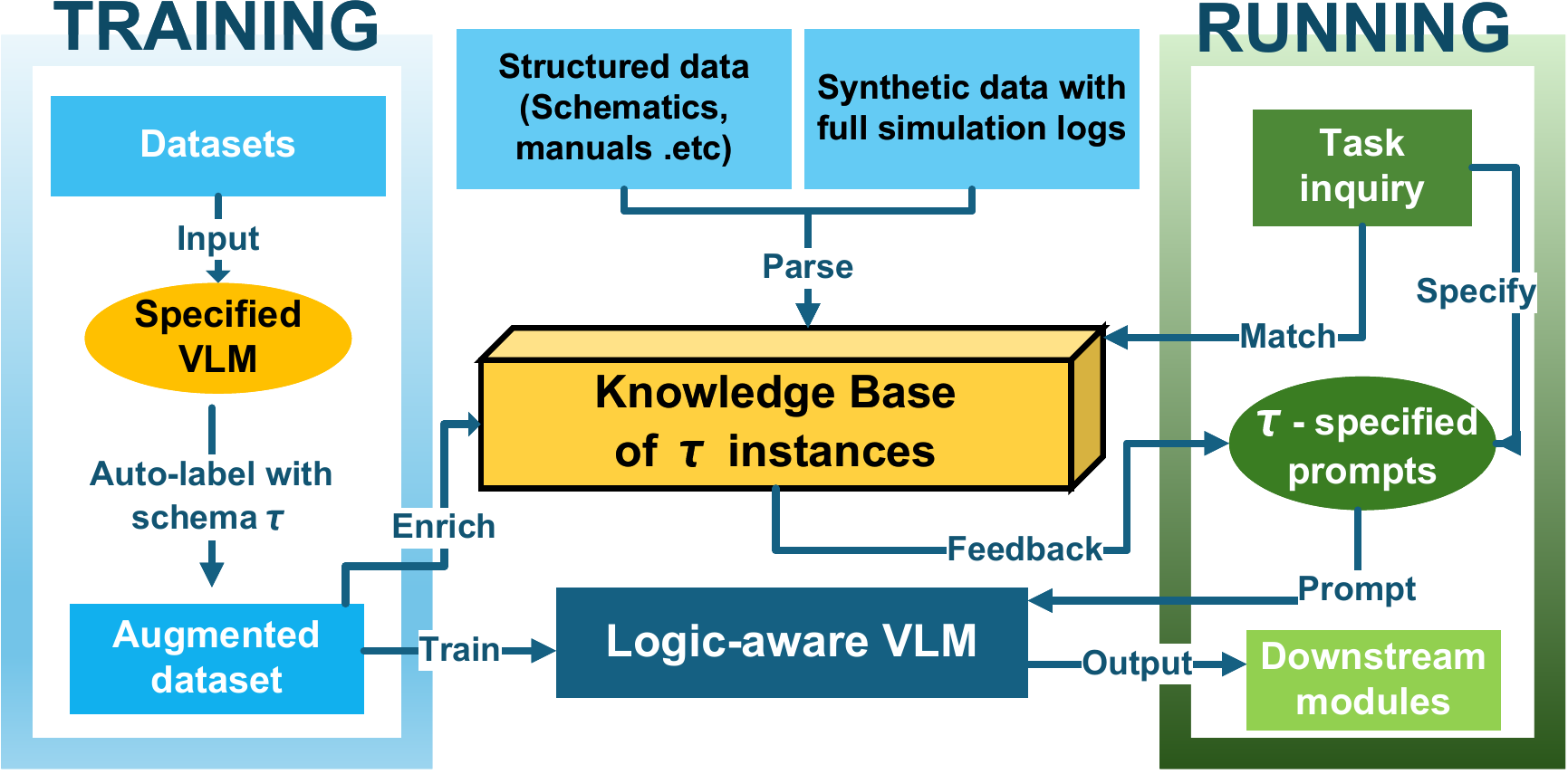}
  \caption{Pipeline of the proposed system design, highlighting the two uses of $\tau$ schema. In training, datasets are auto-labeled with the schema to inject $\tau$ in VLM, also enriching the knowledge base constructed with parsed structured data (schematics, manuals, etc) and synthetic data (with full simulation logs). In running, task inquiry is processed to $\tau$-specified prompt by matching in knowledge base, prompting a $\tau$-conditioned VLM planner module.}
  \label{fig:pipeline}
  \vspace{-2mm}
\end{figure*}

\section{Manipulation Logic schema}\label{sec:schema}

Appearance and generic language omit execution variables decisive for success in contact-rich manufacturing environments: mechanism type, approach constraints, motion direction/timing, precision/tolerance bands, and force/impedance with verification cues. Classical manipulation and contact control codify why these parameters must be explicit for execution \citep{hogan_impedance_1984,raibert_hybrid_1981,mason_mechanics_2001}; manufacturing studies likewise show insertion, fastening, and adjustment succeed only when tolerances and interaction forces are respected \citep{chen_robust_2025,qin_recent_2026}. We lift these signals into an object-centric schema spanning selection, execution, and verification, consistent with recent affordance and part-grounding advances linking parts to operations \citep{tong_oval-prompt_2024,qian_affordancellm_2024,nguyen_open-vocabulary_2023}. The result is a VLM-friendly representation for training augmentation and test-time prompting in collaborative manufacturing cells, where precision, compliance, and safety limits are first-class \citep{zhu_compliant_2025,ISO10218-2-2011}. We serialize each interaction as
\[
  \tau=\langle \texttt{obj}, \texttt{iface}, \texttt{pre}, \texttt{contact}, \texttt{prim}, \texttt{traj}, \texttt{tol}, \texttt{dyn}\rangle.
\]

\begin{enumerate}
  \item \textbf{Object class \& part geometry (\texttt{obj}).} Device taxonomy, target part identifier, salient geometric features, and nominal part/robot frames. \citep{mason_mechanics_2001}

  \item \textbf{Interface mechanism (\texttt{iface}).} Mechanism class (button, knob, latch, lever, hinge, touchscreen, valve) and operation mode (push/rotate/slide); admissible DoF and actuation side. \citep{tong_oval-prompt_2024,nguyen_open-vocabulary_2023}

  \item \textbf{Preconditions \& state (\texttt{pre}).} Interlocks, safety states, tool presence, required orderings and guards for enabling execution. \citep{ahn_as_2023,huang_inner_2023}

  \item \textbf{Contact modality \& constraints (\texttt{contact}).} Intermittent vs sustained contact, alignment, approach vector, and any fixture/compliance assumptions. \citep{hogan_impedance_1984,raibert_hybrid_1981}

  \item \textbf{Motion primitive \& directionality (\texttt{prim}).} Primitive verb and direction (press, pull, slide, lift, twist $\mathrm{cw}/\mathrm{ccw}$); axis unit vector and sign if applicable. \citep{mason_mechanics_2001}

  \item \textbf{Trajectory profile \& timing (\texttt{traj}).} Sequence of phases (e.g., approach, engage, ramp, dwell, sweep, retreat) with parameters and time or event conditions.\citep{chen_robust_2025,qin_recent_2026}

  \item \textbf{Precision \& tolerance bands (\texttt{tol}).} Admissible pose and clearance bands with explicit SI units, e.g., translational $(\delta_x,\delta_y,\delta_z)$ in \si{\milli\metre}, rotational $(\delta_\alpha,\delta_\beta,\delta_\gamma)$ in \si{\degree}; insertion depth/fit classes; allowable force windows when task-specified. \citep{tipary_tolerance_2021}

  \item \textbf{Force/impedance \& feedback cues (\texttt{dyn}).} Split into numeric limits and runtime checks:
    \begin{itemize}
      \item \texttt{dyn.num}: numeric targets/limits (force/torque ranges, stiffness/damping, velocity caps) in SI units.
      \item \texttt{dyn.checks}: runtime tactile/visual checks, success predicates, and abort conditions.
    \end{itemize}
    
Numeric limits originate from specifications, logs, or calibrated procedures, not inferred from vision. \citep{ISO_TS15066_2016}
\end{enumerate}

The elements in $\tau$ are engineered for smart manufacturing: unit-bearing tolerances and force/impedance fields mirror specification sheets and QA limits; interface and part semantics align with BOM/manual vocabulary for retrieval; preconditions and runtime checks implement collaborative safety and guarded moves; provenance and versioning of numeric limits support auditability and change control; and the tuple composes with TAMP and impedance controllers as a checkable execution contract \citep{zhu_compliant_2025,ISO10218-2-2011}. Concretely, \texttt{obj} and \texttt{pre} capture object class and admissible state, \texttt{contact} and \texttt{prim} describe the contact patch and motion primitive, and \texttt{traj}, \texttt{tol}, \texttt{dyn} bind trajectory, admissible error bands, and force/impedance checks into execution variables that can be serialized into prompts and enforced by low-level controllers.


\section{System \& Uses}\label{sec:system}
We use the object-centric manipulation logic schema as the first-class knowledge signal across training and deployment. The schema encodes execution details that appearance and generic instructions omit, and it is the sole interface modules exchange; train and test remain aligned by operating on the same tuple vocabulary. The system sketch is visualized in Fig.~\ref{fig:pipeline}.

\subsection{Knowledge Base and Ingest}
A lightweight store indexes entries by $(\texttt{obj.class},\allowbreak \texttt{obj.part},\allowbreak \texttt{iface.mechanism})$
 and returns a serialized $\tau$ plus provenance. Population paths: (i) manuals and schematics parsed into steps, constraints, and numeric specifications; (ii) instrumented runs providing time-aligned state, force/torque (F/T), and success flags; (iii) guarded calibration that writes discovered bounds to \texttt{dyn}; (iv) staged promotion from train-time tags (non-numeric only; excludes \texttt{dyn.num}). Entries are versioned and carry confidence/expiry. Manuals and datasheets are a primary source for interaction data: we parse procedural steps, device states, and numeric specifications into the tuple so the model reads a structured interaction entry rather than guessing from generic priors \citep{tie_manual2skill_2025}. Retrieval augments this with affordance and part-level constraints at inference, reducing ambiguity when equipment varies \citep{ahn_as_2023,li_improving_2024}.

\subsection{Train-time: tagged augmentation}
Demonstrations are auto-labeled for all non-\texttt{dyn} fields by pairing VLM proposals with rule templates, yielding concise, schema-valid tags (e.g., \textit{obj=valve,part=handle; iface=rotary,mode=twist; prim=twist,cw; traj=ramp+dwell; tol=angle\,$\pm$2$^{\circ}$; dyn=\{checks=click\}}). Tags are fused with images/states via concatenated tokens and/or auxiliary heads so the model learns mappings from visual signatures to the correct manipulation logic rather than memorizing nouns. This structured augmentation is a scoped extension of instruction augmentation and object-aware conditioning shown to improve manipulation generalization and zero-/few-shot transfer \citep{shridhar_cliport_2022,nair_r3m_2023,xiao_robotic_2023,stone_open-world_2023}. Successful tags are staged as knowledge base candidates (non-numeric fields plus documented \texttt{dyn.checks}); \texttt{dyn.num} is never promoted.

\textit{Essential inputs for \texttt{dyn}.} We do not auto-label any \texttt{dyn} field. Numeric interaction parameters \texttt{dyn.num} (e.g., torque bands, stiffness, dwell) are injected from synchronized F/T or joint-torque logs, disturbance-observer estimates, or retrieved manuals/datasheets; when unknown, values are discovered via guarded, impedance-limited calibration under collaborative safety limits \citep{ISO10218-2-2011,zhou_variable_2025,liu_sensorless_2021}. Optional textual \texttt{dyn.checks} (e.g., `click seated,'' no slip'') may be included only when sourced from documentation. This follows compliant/impedance practice where interaction forces are measured, estimated, or bounded, not guessed from images \citep{ISO10218-2-2011,zhou_variable_2025,liu_sensorless_2021}. We envision a training objective that predicts actions and schema fields per phase (approach, engage, verify) with consistency losses that tie language/vision to process parameters.

\begin{table*}[b]
\caption{\NoCaseChange{Logic-aware API}}
\label{tab:api}
\centering
\begin{threeparttable}
\footnotesize
\setlength{\tabcolsep}{6pt}
\begin{tabularx}{\textwidth}{@{}l l X@{}}
\toprule
\textbf{Name} & \textbf{Signature} & \textbf{Semantics} \\
\midrule
lookup &
\(\big(\mathtt{obj.class},\mathtt{obj.part},\mathtt{iface.mechanism}\big)\to(\tau,\ \mathrm{conf},\ \mathrm{prov})\) &
Returns the matching tuple \(\tau\) from a small hand-authored KB, with confidence (1.0 if unique match, 0 if none) and provenance. \\

prompt &
\(\big(\mathrm{context},\ \tau\big)\to \mathrm{prompt\_str}\) &
Compose the planner prompt from \(\tau\) fields and constraints. \\

plan &
\(\big(\mathrm{prompt\_str},\ \mathrm{percepts}\big)\to(\pi,\ \mathrm{cites})\) &
Produce an interaction plan \(\pi\) using an VLM-based planner; \(\pi\) is scored against \(\tau\) during evaluation and is expected to make explicit use of fields \(\{\mathtt{obj,iface,pre,contact,prim,traj,tol,dyn}\}\). \\

execute &
\(\big(\pi,\ \tau\big)\to(\mathrm{status},\ \mathrm{logs})\) &
Run control honoring \(\tau.\mathtt{traj}\); emit checks and aborts per \(\mathtt{dyn.checks}\). \\
\bottomrule
\end{tabularx}
\begin{tablenotes}[para,flushleft]
\footnotesize
\end{tablenotes}
\end{threeparttable}
\end{table*}

\subsection{Runtime: logic-aware prompting/retrieval}

At deployment we query the KB introduced above: perception produces $(\texttt{obj.class}$, $\texttt{obj.part}$,$\texttt{iface.mechanism})$ with uncertainty; a retriever matches these to a KB entry; and a prompt composer provides the VLM with the task context and the full tuple fields. The planner should ground its plan in that tuple, explicitly citing all $\tau$ elements,  with \texttt{dyn.checks} for verification. Control consumes that logic plus \texttt{traj}/\texttt{dyn} and executes with online checks.

When the KB lacks numeric interaction parameters, the system performs a guarded sweep under collaborative safety limits to discover a valid operating range and writes the result back into \texttt{dyn.num} for future use \citep{ISO10218-2-2011}. If observations contradict \texttt{pre}/\texttt{checks} or a numeric bound is missing, the system falls back to conservative defaults, runs the guarded calibration sweep, and commits the updated \texttt{dyn} back to the KB. The contract is minimal: retrieve the matching tuple for the detected part, pass the tuple to the model, require the model's interaction logic to explicitly reference its fields, and let the execution backend enforce the referenced trajectory, tolerance, and interaction limits.

\subsection{Logic-aware API}
See Table~\ref{tab:api} for a minimal demonstration of the logic-aware API. Plans must reference tuple fields, and the back-end enforces trajectory, tolerance, and interaction limits.In our current prototype, the knowledge base is a small table of $\tau$ entries keyed by $(\texttt{obj.class}, \texttt{obj.part}, \texttt{iface.mechanism})$. The \texttt{lookup} call is implemented as an exact dictionary lookup; approximate or graph-based retrieval is left for future work.

\section{Evaluation Protocols}\label{sec:evaluation}

We structure evaluation at three levels: representation, VLM plan quality, and future deployment.

\subsection{Representation-level metrics.}

We use a single representation metric: \emph{logic coverage}, the fraction of execution-critical parameters for an interface that are explicitly populated in $\tau$ fields such as \texttt{pre}, \texttt{traj}, \texttt{tol}, and \texttt{dyn}.

\subsection{Plan-level metrics (instantiated in the case study).}

For the 3D-printer spool-removal case in Sec.~\ref{sec:case-study}, we instantiate non-hardware, plan-level evaluation by sampling VLM plans under different prompting conditions and scoring them against the reference tuple $\tau_{\text{spool\_remove\_discard}}$. We adapt standard VLM/LLM planning metrics: step coverage, order validity, safety and constraint coverage, contact and tolerance specificity, and plan length.

\subsection{Train-time and deployment metrics (future work).}

At training time, we anticipate sample-efficiency and logic-consistency metrics, such as success on unseen but $\tau$-similar interfaces when models are trained with schema-tagged data versus baselines, and the fraction of generated steps that satisfy $\tau$-specified preconditions, contact modes, trajectories, and tolerances. At deployment, execution metrics would include first-try task success, force-limit violations per 100 executions, contact retries per attempt, time-to-completion, and calibration overhead when \texttt{dyn.num} is missing. We do not instantiate these metrics in this paper; they define how we will evaluate $\tau$-conditioned VLM-based controllers once deployed on hardware.

\section{Case Study: Logic-Aware Assistance for Spool Removal}
\label{sec:case-study}

This case study instantiates $\tau$ within the logic-aware API for a 3D-printer spool-removal and discard task.
We do not train a new model; instead, we condition an off-the-shelf multimodal VLM (ChatGPT-4o) with $\tau$-augmented image--text prompts and measure plan quality using the metrics in Sec.~\ref{sec:evaluation}.

\subsection{Scenario and Baseline Workflow}

We consider a desktop fused-filament 3D printer installed in a small production cell, equipped with a hinged top lid, a passive spindle for filament spools, and a waste bin for discarded rolls. A dual-arm mobile operating platform (in our setup we use Airbot MMK2) can reach the printer and the bin, but in the current deployment the empty-spool removal routine is still performed by the human operator.

The routine task is:

\begin{quote}
\emph{``Remove the empty PLA filament spool from the printer and
discard it in the waste bin.''}
\end{quote}

In practice, the operator follows an implicit standard operating procedure (SOP): verify that the printer is idle and cooled, open the
lid, support the spool, slide it axially off the spindle, transfer it to the bin, and retreat. The written SOP, however, typically compresses these details into one or two high-level bullets (for example, ``remove empty spool and discard''), leaving contact, motion direction, tolerances, and safety checks as unstated tacit knowledge.

When a generic vision--language model is prompted only with this high-level instruction, a short textual description of the printer
and cell, and static images from the head and wrist cameras (but no interaction tuple), its responses resemble written SOPs: it
suggests opening the cover, pulling the spool off, and placing it into a bin.

Consistent with failure modes reported in VLM/LLM-based planning and embodied decision making \citep{guo_Open_2024, cao_Large_2025,li_Embodied_2024}, these plans tend to: (i) omit temperature and motion-safety preconditions, (ii) underspecify where and how the spool should be grasped, and (iii) ignore axial constraints, clearances, and force limits that are critical for safe execution on actual robots.

\begin{figure}[t]
  \centering
  \includegraphics[width=7.2cm]{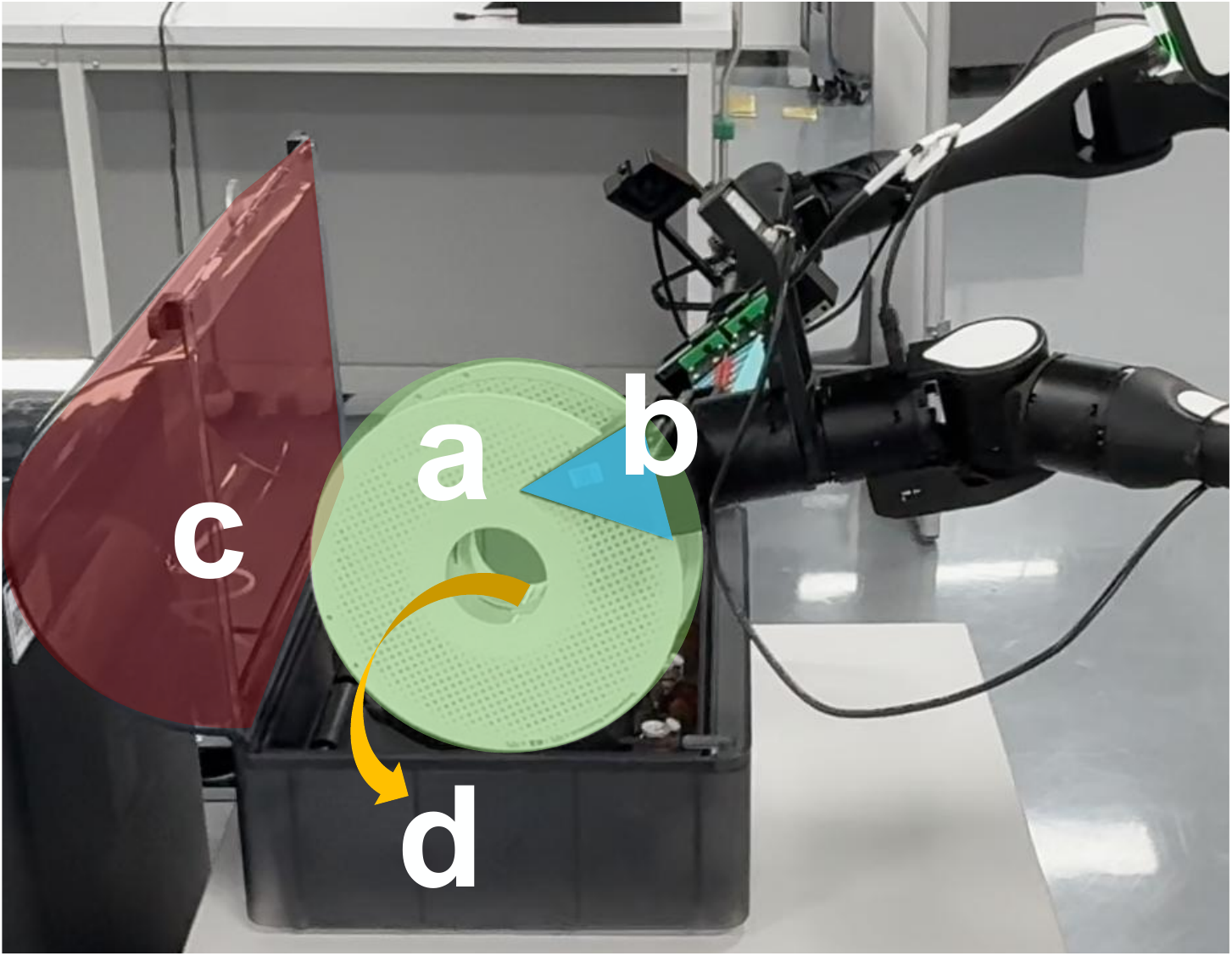}
  \caption{Interaction tuple for the spool-removal task
  $\tau_{\text{spool\_remove\_discard}}$.
  (a) Green mask: filament spool and hub, grounding \texttt{obj}.
  (b) Blue mask: gripper and hub contact, instantiating \texttt{contact} and
  fixing the action axis for \texttt{prim}.
  (c) Red mask: hinged lid, grounding \texttt{iface} and the precondition
  ``lid open'' in \texttt{pre}.
  (d) Yellow arrow: phase motion (approach, open, grasp, extract,
  transfer, deposit), encoding \texttt{traj} and associated primitives
  in \texttt{prim}. Tolerances \texttt{tol} and dynamics \texttt{dyn}
  are defined in the tuple and omitted for clarity.}
  \label{fig:demo}
  \vspace{2mm}
\end{figure}

\subsection{$\tau$-Tuple Instantiation for Spool Removal}
\label{subsec:tau-spool}

To make the interaction logic explicit and machine-readable, we encode
the empty-spool removal task as the object-centric interaction tuple
$\tau_{\text{spool\_remove\_discard}}$ (Fig.~\ref{fig:demo}): 

\medskip
\noindent
\twolinetauB{}{%
\text{obj}=(\text{filament spool}, \ \text{cylindrical core,two flanges})\\
\text{iface}=(\text{hinged lid: rotate-to-open};\\
\qquad \text{spool on passive spindle: axial pull};\ )\\
\text{pre}=(\text{printer idle \& hotend, bed below safe temp},\\ 
\qquad \text{motors disabled or compliant},\ \text{lid unlatched},\\
\qquad \text{spindle bore located},\ \text{bin location verified})\\
\text{contact}=(\text{handle pinch on lid};\ \text{hub pinch on spool};\\
\qquad \text{coaxial align to spindle};\ \text{avoid flange scrap})\\
\text{prim}=(\text{open lid: lift/rotate to }90^{\circ});\\
\qquad \text{grasp spool: pinch};\
\text{extract: axial pull};\\
\qquad \text{transfer};\ \text{lower};\ \text{release};\ \text{retreat})\\
\text{traj}=[(\text{approach lid},\ v=80\,\si{mm/s},\ \text{event: contact})\\
\qquad \rightarrow (\text{open lid},\ \dot\theta=40\,\si{\degree/s},\ \text{event: stop at }90^{\circ})\\
\qquad \rightarrow (\text{approach spool hub},\ v=60\,\si{mm/s},\\
\qquad \qquad \text{event: both fingers seated on hub})\\
\qquad \rightarrow (\text{grasp},\ 6\,\si{N}\ \text{per finger},\ \text{event: no slip})\\
\qquad \rightarrow (\text{extract},\ v_{\text{axial}}=50\,\si{mm/s},\\
\qquad \qquad \text{event: spindle cleared (force drop }>\ 20\%\text{)})\\
\qquad \rightarrow (\text{transfer to bin},\ v=200\,\si{mm/s},\\
\qquad \qquad \text{event: above bin pose reached})\\
\qquad \rightarrow (\text{lower},\ v=80\,\si{mm/s},\ \text{event: top}=80\,\si{mm})\\
\qquad \rightarrow (\text{release},\ \Delta w=+40\,\si{mm}\ \text{gripper width})\\
\qquad \rightarrow (\text{retreat},\ \Delta x=100\,\si{mm})]\\
\text{tol}=(\text{coaxial error }\le 2\,\si{mm};\ \text{roll tilt }|\alpha|\le 5^{\circ};\\
\qquad \text{grip width }[w^\star\!\pm\!3]\,\si{mm};\ \text{pose }(\pm 15\,\si{mm},\ \pm 5^{\circ}))\\
\text{dyn}=(\text{num: }F_{\text{grip}}\in[4,8]\,\si{N}\ ;\ a_{\text{carry}}\le 1\,\si{m/s^2};\\
\qquad \text{speed caps near human zone }v\le 250\,\si{mm/s};\\
\qquad \text{checks: slip}\Rightarrow\text{increase }F_{\text{grip}}\ \text{or abort};\\
\qquad\text{collision/force spike}\Rightarrow\text{stop})%
}
\medskip

In the knowledge base, this instance is stored under the key
$(\texttt{obj.class}=\text{3D\_printer},\,
 \texttt{obj.part}=\text{spool},\,
 \texttt{iface.mechanism}=\text{spindle})$,
so a \texttt{lookup} call with this triple returns
$\tau_{\text{spool\_remove\_discard}}$ with confidence $1.0$ and
hand-authored provenance. This instance binds all eight fields of the
tuple for the specific cell configuration.

The fields in $\tau_{\text{spool\_remove\_discard}}$ are grounded in
three sources: machine documentation (thermal limits, reach envelopes),
cell layout models (spindle and bin poses, clearances), and operator
experience (preferred grasp locations, acceptable misalignments, and
practical force bounds). Several fields (\texttt{pre},
\texttt{tol}, and \texttt{dyn}) are quantitative and safety-critical;
they are not suitable targets for VLM/LLM hallucination and must be provided
by the knowledge base.

\subsection{Minimal Logic-Aware Assistant Prototype}

To examine the effect of exposing this tuple to a VLM, we implement a minimal logic-aware assistant around a VLM planner.
Each query supplies the model with the task instruction, a brief description of the printer and cell, and three RGB frames
(one from the head camera and one from each wrist camera) capturing the scene.
The assistant maintains a small knowledge base and uses \texttt{lookup} on the object/interface triple and task identifier to retrieve $\tau$.
It then renders the result into a structured prompt segment that cites all fields in the tuple.

We then compare two prompting conditions for the same backbone VLM:

\begin{itemize}
    \item \textbf{Baseline (no-$\tau$):} the model receives the natural-language instruction, the brief description of the printer and cell, and the three camera images; it is asked to propose a numbered list of robot actions.

    \item \textbf{$\tau$-anchored (logic-aware):} the model receives the same text and images, plus the rendered tuple segment. The prompt explicitly instructs the model to respect all listed preconditions and constraints, and to map each step back to tuple fields where possible.
\end{itemize}

For each condition, we sample $N$ plans (here $N{=}10$) by repeating the query with fixed decoding parameters. Following recent work on VLM/LLM planning evaluation \citep{guo_Open_2024,cao_Large_2025,yang_Plug_2024,li_Embodied_2024}, we treat $\tau_{\text{spool\_remove\_discard}}$ as a reference specification and score each plan using:

\begin{itemize}
    \item \textbf{Step coverage (completeness).}
    Fraction of primitive and precondition items from the tuple that appear as explicit steps or clauses in the plan, analogous to completeness metrics in Open Grounded Planning \citep{guo_Open_2024}.

    \item \textbf{Order validity.}
    Proportion of plans that respect the partial order implied by the tuple (safety checks before opening the lid, lid open before grasp, spindle cleared before transfer, retreat after deposit), following temporal-order checks in VLM/LLM planning surveys \citep{cao_Large_2025}.

    \item \textbf{Safety and constraint coverage.}
    Proportion of plans that explicitly mention thermal safety, speed caps in shared zones, and slip/collision checks derived from \texttt{pre}, \texttt{tol}, and \texttt{dyn}, mirroring the separation of success rate and safety rate in constrained LLM agents \citep{yang_Plug_2024}.

    \item \textbf{Contact and tolerance specificity.}
    Average fraction of manipulation steps that specify contact locations, approach directions, grip widths, or allowable misalignments, similar to parameter-grounding metrics for embodied decision making \citep{li_Embodied_2024}.

    \item \textbf{Plan length.}
    Average number of steps per plan, a coarse proxy for efficiency and redundancy used in several VLM/LLM planning benchmarks \citep{guo_Open_2024,cao_Large_2025}.
\end{itemize}

The resulting comparison, averaged over the $N{=}10$ samples per condition, is summarized in Table~\ref{tab:case-study}. The absolute values depend on the particular model and prompt template; here they are representative of what we observed for a modern general-purpose VLM.

\begin{table}[b]
  \centering
  \caption{Plan quality metrics for the filament spool removal case study}
  \label{tab:case-study}
  \setlength{\tabcolsep}{4pt}
  \begin{threeparttable}
    \begin{tabular}{lcc}
      \toprule
      \textbf{Metric} &
      \textbf{Baseline} &
      \textbf{$\tau$-anchored} \\
      \midrule
      Step coverage (\%)                  & 68 & 94 \\
      Order validity (\%)              & 50 & 92 \\
      Safety/constraint coverage (\%)     & 66 & 88 \\
      Contact/tolerance specificity (\%)  & 35 & 89 \\
      Avg.\ steps per plan                & 5.3 & 7.6 \\
      \bottomrule
    \end{tabular}
\begin{tablenotes}[para,flushleft]
  \footnotesize
  \item Coverage, order validity, and safety/constraint coverage are reported as percentages; contact/tolerance specificity is the fraction of manipulation steps with explicit parameterization. All values are averages over $N{=}10$ sampled plans per condition, scored using the metrics defined above with $\tau_{\text{spool\_remove\_discard}}$ as the reference specification.
\end{tablenotes}

  \end{threeparttable}
  \vspace{-2mm}
\end{table}

\subsection{Comparison and Discussion}

The case study shows that, even in a single representative scenario, the interaction tuple changes the behaviour of a strong VLM in ways that align with standard plan-quality criteria in the VLM/LLM planning literature \citep{guo_Open_2024,cao_Large_2025,yang_Plug_2024,li_Embodied_2024}. Under instruction-only prompting, the model often produces short plans that omit preconditions, blur contacts (``grab the spool'' without specifying the hub), and ignore axial and safety constraints. When the same model is exposed to the serialized $\tau_{\text{spool\_remove\_discard}}$ and instructed to respect it, the plans show higher coverage of primitives and preconditions, fewer ordering violations, and much more explicit contact and tolerance parameterization. The average plan length increases modestly (from $5.3$ to $7.6$ steps), reflecting that $\tau$-anchored prompts push the model to spell out safety checks and contact details instead of compressing them into one vague instruction.

For human--robot collaboration in manufacturing, this matters in three ways. First, the tuple turns tacit manipulation knowledge into a structured contract that can be checked and audited independently of any particular VLM. Second, the gains in step coverage, ordering, and safety coverage indicate that even a simple retrieval-plus-prompting layer can move generic VLM planning outputs closer to the requirements of collaborative cells and industrial safety standards. Third, the case study shows that the taxonomy is not only descriptive: it can be instantiated in a realistic task and used to organize the knowledge base and the evaluation of VLM/LLM-generated manipulation plans.

\section{Limitations and Outlook}\label{sec:limits}

Our approach assumes that the retrieved tuple matches the actual interaction, that perception correctly grounds objects and interfaces, and that KB entries remain up to date. Key open limits include deformable objects, simultaneous multi-contact, and incomplete KB coverage. Once $\tau$-conditioned controllers are deployed on hardware, we plan to instantiate the execution-time metrics in Sec.~\ref{sec:evaluation}.

Looking forward, manufacturing is unusually well suited to a shared interaction-logic ecosystem: objects are structural, reused, and documented. This supports a public, queryable interaction-logic KB built from manuals, GD\&T/process specifications, and instrumented runs, so models read device-specific procedures instead of extrapolating from generic priors; over time, that KB can supervise training of logic-conditioned VLM planner modules specialized for manufacturing \citep{sliwowski_demonstrating_2025}. Beyond the prompt-only use in this paper, a natural next step is to $\tau$-tag demonstration logs and benchmarks such as RLBench and finetune VLM-based planners so that the plan-quality metrics in Sec.~\ref{sec:evaluation} (completeness, constraint and safety coverage, parameter specificity) improve out of the box on unseen interfaces. Numeric interaction limits drawn from standards and vendor specifications, rather than model guesses, make the approach explicitly anti-hallucination and compatible with collaborative-safety practice. Because the tuple and API are orthogonal to policy class (diffusion, transformer, or planner hybrids), they remain usable as VLA trends evolve, and Manual2Skill- and FOON-style pipelines suggest how the KB can grow beyond a hand-authored dictionary into a graph that supports approximate retrieval and generalization \citep{tie_manual2skill_2025,paulius_functional_2018}.

\bibliography{references_final_clean}             
                                                   







\end{document}